\documentclass[11pt]{article}
\usepackage{coling2020}
\usepackage{times}
\usepackage{url}
\usepackage{latexsym}
\usepackage{epsfig}
\usepackage{graphicx}
\usepackage{multirow}
\usepackage{multicol}
\usepackage{subfigure}
\usepackage{booktabs}
\usepackage{hyperref}
\usepackage{amsmath}
\usepackage{amssymb}
\usepackage{multirow}
\usepackage{multicol}
\usepackage{subfigure}

\usepackage{color}
\usepackage{framed}
\usepackage{graphicx}
\usepackage{url}
\usepackage{times}
\usepackage{epsfig}
\usepackage{latexsym}
\usepackage{graphicx}
\usepackage{multirow}
\usepackage{multicol}
\usepackage{xcolor,colortbl}
\usepackage{booktabs}
\usepackage{amssymb}
\usepackage{subfigure}
\usepackage{hyperref}

\definecolor{green}{RGB}{127,255,0}
\definecolor{limegreen}{RGB}{127,255,0}

\colingfinalcopy 

\title{Enhancing Fine-grained Sentiment Classification Exploiting Local Context Embedding}

\author{
	Heng Yang \\
	School of Computer, \\
	SCNU, China\\
	{\tt yangheng@m.scnu.edu.cn} \\\And
	Biqing Zeng \\
	School of Software, \\
	SCNU, China\\
	{\tt zengbiqing@scnu.edu.cn}
}

\date{}

\begin{document}
	\maketitle
	\begin{abstract}
	Target-oriented sentiment classification is a fine-grained task of natural language processing to analyze the sentiment polarity of the targets. To improve the performance of targeted sentiment classification, existing methods mostly are based on traditional attention mechanisms to capture the information within important context words. Meanwhile, the local context focus (LCF) mechanism provides a novel approach to capture the significant relatedness of a target's sentiment and its local context, achieving considerable performance. However, the original LCF does not notice the extra enhancement of local context embedding (LCE). In this paper, we propose a enhanced local context-aware network (LCA-Net) based on local context embedding. Moreover, accompanied by the sentiment prediction loss, the local context prediction (LCP) loss is proposed to make full use of the LCE. We implement the LCA-Net with different neural networks, and the experimental results on three common datasets, the Laptop and Restaurant datasets from SemEval-2014 as well as a Twitter social dataset, show that all the LCA-Net variants perform superiorly to existing approaches in extracting local context features.

	\end{abstract}
	
	\section{Introduction}
	\label{intro}
	
	Target-oriented sentiment classification (TSC) is a sophisticated subtask of sentiment analysis \cite{liu2012sentiment,bakshi2016opinion,cambria2016affective}, aiming to infer the sentiment polarities (e.g. positive, negative and neutral) of recognized targets (a.k.a. aspect-based sentiment classification \cite{pontiki2014semeval,pontiki2015semeval,pontiki2016semeval}, ABSC). Given a customer's review ``\textit{The \textbf{screen resolution} attracts me but its \textbf{battery} is miserable}'', the sentiment polarity of ``\textbf{\textit{screen resolution}}'' is positive while the polarity of ``\textbf{\textit{battery}}'' is negative.
	
	Recent research tended to adapt the recurrent neural networks (RNNs) or convolutional neural networks (CNNs) to solve TSC instead of using machine learning or rule-based methods. Those methods regard the target as important supplemental information to infer the polarities. The traditional attention mechanism introduced in machine translation \cite{bahdanau2014neural} has been adapted to TSC by some approaches. For example, Wang et al. \cite{wang2016attention}, Ma et al. \cite{ma2017interactive} and Fan et al. \cite{fan2018multi} proposed the attention-based models to learn the semantic and sentiment relatedness between target and context words.
	
	\begin{figure*}[h]
		\centering
		\includegraphics[width=1\columnwidth]{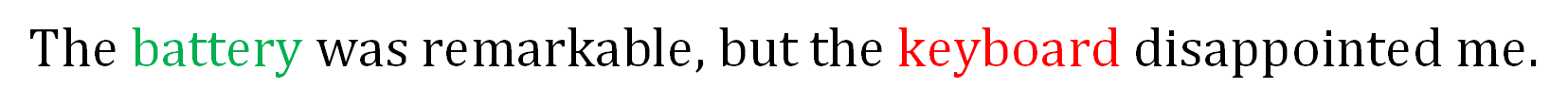}
		\caption{An example review of targeted sentiment analysis about laptops. The ``battery'' and ``keyboard'' are the targets with fine-grained sentiment polarities of \textit{positive} and \textit{negative}.}
		\label{fig:sample}
	\end{figure*}
	
	However, the improvements of classical attention techniques for TSC are limited, because deploying multiple sophisticated attentions increases computation and reduces inferring efficiency.
	Meanwhile, RNN-based frameworks are insufficient to learn the semantic features of remote context words of a target: while a word is far away from the target in the context, it's hard to precisely extract the correlation between the target and the context. Therefore, many studies explore to eschew RNNs when building models. Self-attention is a novel attention mechanism with a stronger semantic feature extraction ability. And the self-attention \cite{vaswani2017attention} is effectively to extract the semantic relatedness in remote context words based on parallel matrix calculation. The importance of local context focus in TSC has been proved in recent works \cite{yang2019multitask,phan2020modelling,luo2020does}. The empirical observation is that the context words which are neighbor to a target are more semantic-relevant to the target. In that case, more sentiment information is possibly contained in the target's local context rather than the remote context.
	
	However, the original LCF failed to utilize the LCE to model for local context extraction by the LCP loss. Besides, the LCF-BERT \cite{zeng2019lcf} is resource-consuming and too large to train.
	In this paper, we propose the local context aware network (namely LCA-Net) \footnote{The codes for this paper are available at \url{https://github.com/yangheng95/LC-ABSA}.} and studied the significance of local context embedding and local context prediction loss, enhancing the model to capture local context semantic features from the input text. Besides, we compress the LCF design by using the same BERT-shared layer to extract local context features and global context features. Apart from implementing the baseline model based on the self-attention, we also implement an enhanced version (a.k.a. LCA-BERT) integrating with BERT (Bidirectional Encoder Representations from Transformers) \cite{devlin2019bert}. Experimental results on three commonly used TSC datasets show that the local context embedding and local context prediction loss can significantly improve the performance of TSC, which provides a guideline for using local context to improve other target-level natural language processing tasks.
	
	The main contributions of this paper are as follows:
	\begin{enumerate}
		\item Two implementations of local context embedding (LCE) layer and the local context prediction (LCP) are proposed to make full use of local context words and enhance the performance of the LCA-Net.
		\item  We conduct experiments on commonly used datasets to develop the different embedding methods of local context by implementing the LCA-Net using various neural networks, which valid the effectiveness and scalability of LCA architecture.
		\item The evaluation of the effectiveness of LCE and LCP loss via ablation studies indicates the LCE and LCP are highly adaptive that can strengthen other TSC models.
	\end{enumerate}

	\section{Related Works}
	
	The TSC was regarded as a fine-grained text classification task in previous studies. Traditional machine learning-based approaches \cite{kiritchenko2014nrc,wagner2014dcu,vo2015target} generally relies on manually designed features and lexicon features, etc., which are inefficient and easily reach the performance bottleneck. Neural networks have been proved to be competent for extracting text features and semantic relatedness. Consequently, there are flourishing studies of TSC based on deep neural networks.
	
	\subsection{Traditional Deep Neural Networks based Methods}
	Tang et al. \cite{tang2016effective} proposed TD-LSTM to model the features of the left context and right context of the targets independently, combining these features to analyze polarity. However, the targets are not considered while modeling for the contexts, resulting in potential loss of the sentiment information of the targets itself.
	To exploit the potential information of targets, Wang et al. \cite{wang2016attention} and Ma et al. \cite{ma2017interactive} adopted the attention mechanism to help model focus on the important words to the target. To overcome the obstacles of classical coarse-grained attention, MGAN \cite{fan2018multi} proposed a fine-grained attention mechanism to link and fuse information from the target and the context words. Combining with the coarse-grained and fine-grained attention, MGAN is a multi-attention network and significantly outperforms the coarse-grained attention-based models. 
	For the sentences containing multiple targets, RAM \cite{chen2017recurrent} and TNet \cite{li2018transformation} considers the word position while extracting and learning the features of the context, which alleviates the mutual interference between the contextual sentiment information of multiple targets.
	
	Mao et al. \cite{mao2019aspect} introduced the ANTM that can model the  relatedness between targets and  context. The ANTM can concentrate on important sentiment information by swift IO operations. Du et al. \cite{du2019capsule} noticed that previous research failed to dynamicly learn overlapped features in the context. They proposed the model based on a capsule network which can extract vector-based features and cluster features by the expectation-maximum routing algorithm. Integrated by the interactive attention, the model can learn semantic relationship between targets and context. 
	Sun et al. \cite{sun2019aspect} proposed a model based on LSTM and GCN, which utilizes the dependency trees to learn the semantic correlation between targets.
	
	Lin et al. \cite{lin2019deep} presented a deep mask memory network that involved semantic dependency and context moment, making use of the semantic-parsing information of targets and the semantic relatedness of multi-targets. Liu et al. \cite{liu2020aspect} introduced the GANN based gated alternate neural network which attempted to address the learning of long-distance dependency and modeling sequence information. Zhao et al. \cite{zhao2020modeling} proposed a graph convolutional networks (GCN) based model for TSC that can extract the sentiment dependencies in multi-targets scenarios. Huang et al. \cite{huang2020aspect} proposed the joint topic embeddings for target and sentiment in the word embedding space, and adopted few keywords describing each aspect-sentiment pair without using any labeled examples in weakly-supervised training. Shuang et al. \cite{shuang2020feature} proposed a distillation mechanism for text features to reduce noise and distill sentiment features of targets, and introduced the double-gate mechanism to measure the relatedness between context and targets. Chen et al. \cite{chen2020relation} presented a relation-aware model using collaborative learning and multi-task learning for TSC, utilizing the relation-dependent propagation during training process.
	
	\subsection{External Knowledge Enhanced Methods}
	
	In order to further improve the performance of TSC, various external knowledge \cite{young2017augmenting,rietzler2019adapt,zhang2020knowledge,zhou2020sk} is used to enhance the ABSC model, and the experimental results prove the feasibility of knowledge enhancement. Deep contextualized pre-trained language model (e.g. BERT \cite{devlin2019bert}, ALBERT \cite{lan2019albert}), is another implementation of external knowledge enhancement, that can significantly improve the performance of most NLP tasks. Sun et al. \cite{sun2019utilizing} proposed to exploit BERT for a sentence-pair classification model by constructing auxiliary sentences to mine the deep relatedness between aspect and context, and achieved a new state-of-the-art performance. Inspired by aforementioned work, BERT-SPC \cite{song2019targeted} concatenates aspect to sentences as auxiliary information, and significantly improved the sentiment classification performance on the Restaurant dataset and the Laptop dataset \footnote{Both datasets are available at \url{http://alt.qcri.org/semeval2014/task4}.} by construct the as input ``\textit{[CLS]+ sentence +[SEP]+aspect+[SEP]}''.
	
	Existing studies have shown that the distance between context and target aspect affects the association degree \cite{chen2017recurrent,li2018transformation,liu2020aspect} of semantic features. To model the relatedness between semantic features of local context and sentiment of aspects, LCF-BERT \cite{zeng2019lcf} and LCFS-BERT \cite{phan2020modeling} extracts local and global context features of aspects respectively and infer sentiment polarity by interactively learning both local and global context features. Compared with the traditional attention mechanisms, the local context focus is more sensitive to the local semantic features of the aspect, and the improvement of the LCF-based model is significant. The LCA-Net further improves the model's ability to extract local context features by applying local context embedding.
	
	There are growing studies that aim to improve the performance of TSC by combining the pre-trained language models (e.g. ALBERT \cite{lan2019albert}). BERT-PT and AEN-BERT adapt BERT to solve the TSC and improve its performance.

	\section{Local Context-Aware Network}
	
	\subsection{Task Definition}
	
\begin{figure}[t]
		\centering
		\includegraphics[width=0.628\columnwidth]{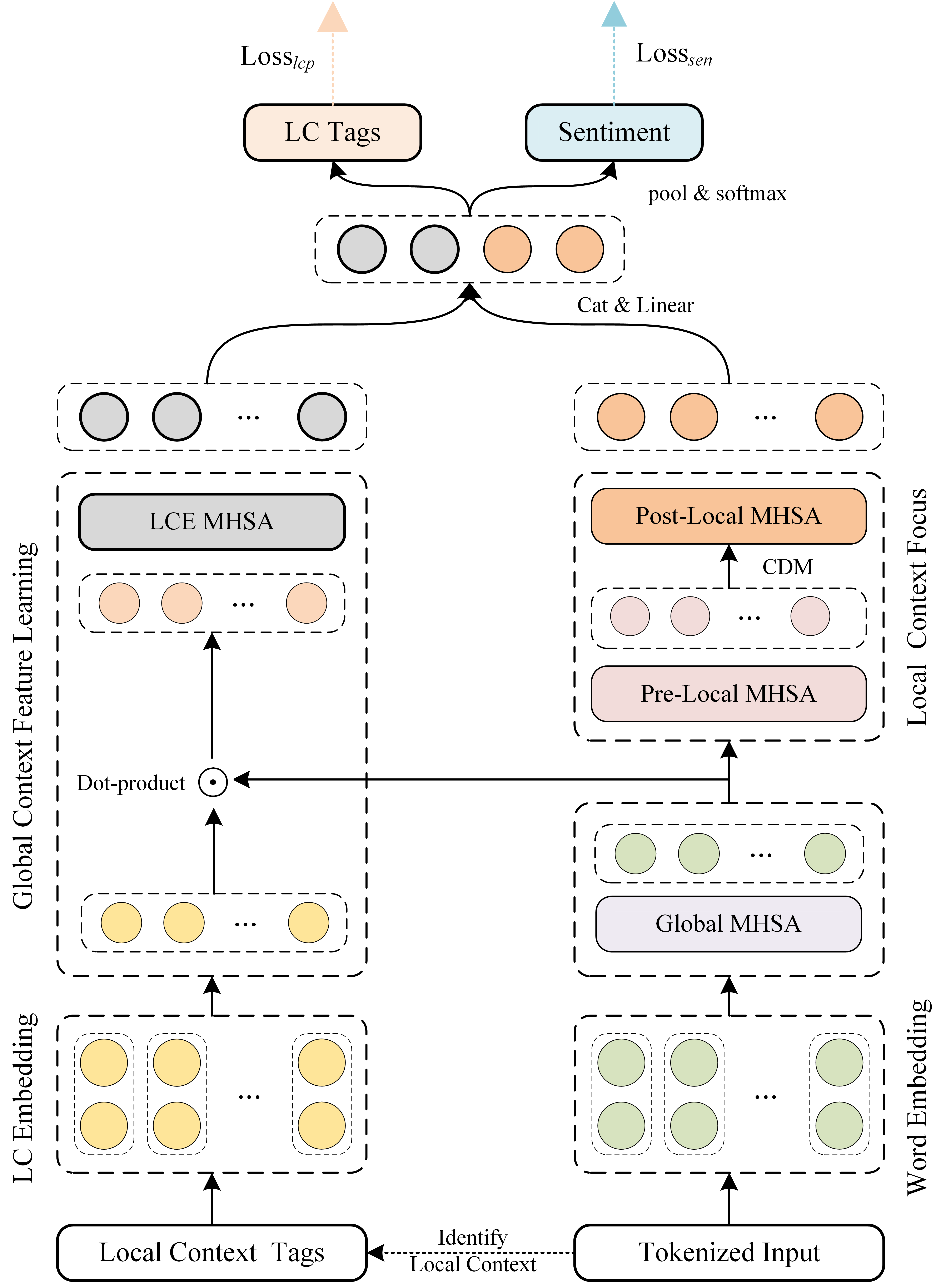}
		\caption{The main architecture of theLCA-Net.}
		\label{fig:lca-net}
	\end{figure}

	Given a sentence ${s}=\left\{w_{0}, w_{1}, \ldots, w_{n}\right\}$ that contains $n$ words including the targets\footnote{To obtain the same dimension of word representations, we conduct truncating and padding for the sentence.}, ${s}^{t}=\left\{w_{0}^{t},  w_{1}^{t}, \ldots, w_{m}^{t}\right\}$ is the target sequence composed of $m~(m\geq1)$ words, and $s^t$ is a subsequence from $s$, and there could be multiple targets in $s$. 
	Fig. \ref{fig:lca-net} is the framework of the local context-aware network.
	
	\subsection{Word Embedding}
	The LCA-Net generates the word representations by GloVe \cite{pennington2014glove}, which maps each word to a vector space. The embedding lookup matrix is denoted as $\mathbb{\textit{E}}^{w} \in \mathbb{R}^{d_{v} \times|V|}$, where $d_{v}$ is the embedding dimension and $|V|$ is the vocabulary size. And we obtain the global context features by applying multi-head self-attention (MHSA) to learn the raw word representations. The pretrained BERT is an alternative for generating word representations in the LCA-Net.
	The global context feature $O^g$ is encoded from the embedded global context $X^g$ by the Global MHSA.
	\subsection{Multi-Head Self-Attention}
	To encode context features, we adopt MHSA which performs multiple scaled dot-product attention (i.e. Attention) in parallel. The MHSA can alleviate potential features loss of the targets' remote context words.
	Suppose $X$ is the feature representation of input sentences, $K$, $Q$, $V$ are the matrices packed from $X$ by multiplying  $W_{q} \in \mathbb{R}^{d_{h} \times d_{q}}$, $W_{k} \in \mathbb{R}^{d_{h} \times d_{k}}$, $W_{v} \in \mathbb{R}^{d_{h} \times d_{v}}$ , where $d_{h}$ is the dimension of the hidden size, and $d_{q}=d_{k}=d_{v}=\sqrt{d_h}$. The attention is calculated as follows:
	\begin{equation}
		\mathrm{Attention(Q,K,V)}=Softmax\left(\frac{Q K^{T}}{\sqrt{d_{k}}}\right)V 
	\end{equation}
	Then we apply MHSA operation on assembled scaled-dot attentions:
	\begin{equation}
		MHSA(X)=\tanh\left(\left\{H_{1};\ldots;H_{h}\right\} W^{o}\right)
	\end{equation}
	where ``;'' denotes vector concatenation, $H$ is the output of each attention head, $h$ is the number of attention heads, and $W^{o} \in \mathbb{R}^{hd_{v} \times d_{h}}$ is the projection matrices.  Besides, we deploy a $\tanh$ activation function for the output of MHSA.
	
	\subsection{Local Context-Aware} 
	The framework of the LCA-Net is composed of the local context embedding, the local context prediction loss, and the local context focus mechanism.
	The local context is identified according to the semantic relative distance (SRD) threshold ($\alpha$), which is proposed to depict the distance between a context word and the target. The SRD $(d_{i})$ of the \textit{i}-th context word relative to a target is calculated as:
	\begin{equation}
		d_{i} = |i-p^{t}|-\lfloor\frac{m}{2}\rfloor
	\end{equation}
	where $i~(1\leq i \leq n)$ denotes the position of the context word,  $p^{t}$ is the average position of the target since a target may contain multiple words, and $m$ denotes the length of target.
	
	\subsubsection{Local Context Embedding}
	We propose a novel local context embedding to enhance the model utilizing the local context features based on the local context tag (LC-tag). The LC-tag flags whether a context word belongs to the local context. For a sentence is ${T}=\left\{T_{0}, T_{1}, \ldots, T_{n}\right\}$, we obtain the LC-tag of the \textit{i}-th context word:
	\begin{equation}T_{i}=\left\{\begin{array}{ll}
			1, & d_{i} \leq \alpha \\
			0, & d_{i}>\alpha
		\end{array}\right.\end{equation}
	The LC-tags are fed into the LCA-Net through a local context embedding matrix $\mathbb{\textit{E}}^{lce} \in \mathbb{R}^{d_{v} \times 2}$. We apply a dot-product operation between the embedded LC-tags $X^{t}$ and the global context features $O^g$ learned by the global MHSA:
	\begin{equation}
		O^{g}_{lce}=X^{t} \odot O^{g}
	\end{equation}
	We also tried the feature-concatenation between $X^{tag}$ and $O^g$. However, it seems that the dot-production plays a better role in exploiting local context embedding according to our experimental results (e.g. the performance of LCF-BERT on the Laptop dataset drops approximately 1.5\% using vector concatenation between $X^{t}$ and $O^{g}$.). It is speculated that the local context embedding is position-wisely associated with the vector of global context features, and can adjust the contribution of the corresponding position's feature.
	
	\begin{figure*}[h]
		\centering
		\includegraphics[width=\columnwidth]{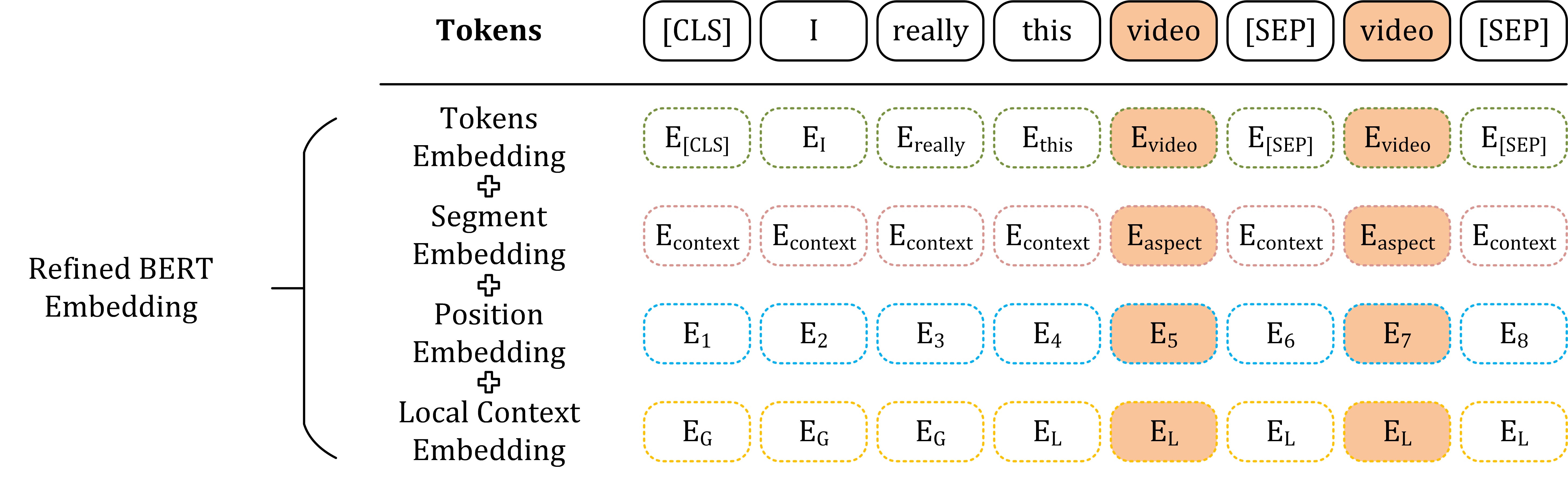}
		\caption{The refined embedding of BERT exploiting LCE. $E_G$ and $E_L$ mean the embedding of local context tag and global context tag, respectively.}
		\label{fig:lce2}
	\end{figure*}
	
	Apart from the implementation of LCE in Fig. \ref{fig:lca-net}, BERT combines the token embedding, the positional embedding as well as  the segment embedding as the input, whihc inspires us to design a new way of using local context embedding. We embed the local context tags into BERT's embedding to improve the learning and prediction performance of local context tags. This implementation of LCE is LCA-Net-E, see Fig. \ref{fig:lce2}.

	\subsubsection{Local Context Prediction Loss}
	
	The motivation behind designing local context-tags prediction loss is that if we can embed LC-tags as auxiliary information so that the model can make use of local context features, then we can also deploy an LCP layer to predict LC-tags. An empirical hypothesis is that if the model can precisely predict LC-tags for local context, the model's ability to extract and learn local contextual features will presumably be enhanced, consequently. It can be seen that LCP is a token-level classification task. The LCP loss is defined as follows:
	
	\begin{equation}
		\mathcal{L}_{lcp}=-\sum_{1}^{N} \sum_{1}^{k} \hat{t}_{i} \log t_{i}
	\end{equation}
	where $N=2$ are the types of LC-tags, $t_i \in \{0,1\}$ are the predicted LC-tags, and $k$ denotes the sum of the context words. 
	
	\subsubsection{Local Context Focus}
	The local context focus (LCF) \cite{zeng2019lcf} mechanism was proposed to extract the local context's features. Compared with classical attention, the LCF reduces the interference of sentiment information contained in multiple target's local contexts and significantly improves the TSC performance. 
	We adopt the context-feature dynamic masking (i.e. CDM) to obtain local context features, which means only the features generated at the local context words' position will be preserved. 
	We mask the features of non-local context words by setting their feature-vectors to zero vectors. The mask vector $V_{i}$ of the  \textit{i}-th context word is generated by:
	\begin{equation}
		V_{i}=\left\{\begin{array}{ll}{E}, & {d_{i} \leq \alpha} \\ {O}, & {d_{i}>\alpha}\end{array}\right.
	\end{equation}
	where $\alpha$ denotes the SRD threshold, $M$ is the mask matrix which contains the mask vector for each word. $E \in \mathbb{R}^{d_{h}}$ is the all-ones vector and $O \in \mathbb{R}^{d_{h}}$ is the all-zeros vector. 
	\begin{equation}
		M=\left[V_{1}, V_{2}, \ldots V_{n} \right]
	\end{equation}
	The local context features $O^l$ are obtained by:
	\begin{equation}
		O^{l}=O_{g} \odot M
	\end{equation}
	To rebalance the  feature-distribution and learn the inner semantic correlation of local context features, a Post-Local MHSA is deployed to learn the local context features $O^{l}$.
	\subsection{Output Layer}
	The LCA-Net concatenates the $O^{g}_{lce}$ and $O^{l}$ and employs linear projections to the global context features and local context features. Then, we take the projected features $O^{p}$ and the first hidden state $O^{head}$ to predict the LC-tags and polarity.
	\begin{equation}
		\hat{T} = softmax(W^{t} O^{p} + b^{t})
	\end{equation}
	\begin{equation}
		\hat{Y} = softmax(W^{y} O^{p} + b^{y})
	\end{equation}
	where $C$ is the number of polarity categories, $ W^{t} \in \mathbb{R}^{N \times d_{h}}$, $ b^{t} \in \mathbb{R}^{N}$, $ b^{t} \in \mathbb{R}^{C}$, $W^{y} \in \mathbb{R}^{C\times d_{h} }$, $b^{y} \in \mathbb{R}^{C}$ are the weight vectors and bias vectors. And $\hat{T}$, $\hat{Y}$ are the predicted LC-tags and sentiment polarities, respectively.
	\subsection{Model Training}
	We use the cross-entropy funciotn as the loss of our polarity classification. We employ the LCP loss and sentiment classification loss to optimize our models. The joint loss function is  calculated by:
	\begin{equation}
		\mathcal{L}=-(1-\sigma)\sum_{1}^{C} \widehat{y_{i}} \log y_{i} - \sigma \mathcal{L}_{lcp} + \lambda \sum_{\theta \in \Theta} \theta^{2}
	\end{equation}
	where $ \sigma \geq 0$ adjusts the influence of $\mathcal{L}_{lcp}$, $\lambda > 0$ is the $L_{2}$ regularization term, and $\Theta$ denotes the parameter set. The optimizer in the LCA-Net is \textit{Adam}. The LCA-Net is implemented on different architectures, such as LSTM, MHSA, and BERT\footnote{For a fair comparison of the improvement of the LCA-Net, the basic BERT was adopted to build LCA-BERT. We implement our models based on \url{https://github.com/huggingface/transformers}. And all the experiments are conducted on the RTX 2080 GPU.}, respectively. We also adapt the domain-adapted BERT\footnote{There is no domain-adapted BERT for the Twitter dataset, we employ the Restaurant domain-adapted BERT, instead.} \cite{rietzler2019adapt} and BERT-SPC \cite{song2019targeted} as the tricks to enhance the LCA-Net. 
	\section{Experiments}
	
	\subsection{Datasets and Hyperparameter Settings}
	To prove the effectiveness of the LCA-Net, we conduct sufficient experiments on three TSC datasets: Laptop, Restaurant and Twitter. The Laptop and Restaurant datasets are obtained from the SemEval-2014 task4\footnote{The datasets can be found at \url{http://alt.qcri.org/semeval2014/task4}.} \cite{pontiki2014semeval}, and the Twitter dataset was presented by \cite{dong2014adaptive}. Table \ref{tab:datasets} and Table \ref{tab:hyperparameters} depicts the details of three datasets and hyperparameters' setting, respectively. Polarities of the targets in these dataset are catergorized into \textit{neutral}, \textit{positive} and \textit{negative}. We employ the \textit{Accuracy} and \textit{macro F1} as the  evaluation metrics to evaluate the performance of our models.
	
	\begin{table}[h]
		\small
		\centering
		\caption{The details of three English TSC datasets.}
		\begin{tabular}{ccccccc}
			\toprule
			\multirow{2}{*}{\textbf{Datasets}}&
			\multicolumn{2}{c}{\textbf{Positive}}&\multicolumn{2}{c}{\textbf{Negative}}&\multicolumn{2}{c}{\textbf{Neural}}\cr
			\cmidrule(lr){2-3}\cmidrule(lr){4-5}\cmidrule(lr){6-7}
			&\textbf{Train}&\textbf{Test}&\textbf{Train}&\textbf{Test}&\textbf{Train}&\textbf{Test} \cr
			\midrule
			Laptop     &994   &341    &870    &128    &463    &169    \cr
			Restaurant  &2164  &728    &807    &196    &631    &196    \cr
			Twitter    &1561  &173    &1560   &173    &3126   &345    \cr
			\bottomrule
		\end{tabular}
		\label{tab:datasets}
	\end{table}

	\begin{table}[h]
		\small
		\centering
		\caption{The details of the hyperparameters of the LCA-Net. All the sentences are padded to a unique length, and is referred to the "padding length". The ``5, 3, 5'' contains the $\alpha$ value for Laptop, Restaurant, and Twitter datasets, respectively. }
		
		\begin{tabular}{ccc}
			\toprule
			\textbf{Hyperparameters}& \textbf{LCA-MHSA} & \textbf{LCA-BERT}\cr
			\midrule
			learning rate   &$2\times e^{-3}$	&$2\times e^{-5}$ \cr
			batch size		&32 	&16	\cr
			hidden size ($d_h$)		&300	&768		\cr
			dropout			&0.1	&0.1	\cr
			training epoch	&10	    &5	\cr
			padding length  &80		&80	\cr
			$h$        		&30	    &12  \cr	
			$\alpha$        &5, 3, 5	    &5, 3, 5  \cr	
			$\sigma$        &0.5	&0.5  \cr	
			$\lambda$       &$1\times e^{-4}$ &$1\times e^{-5}$ \cr	
			\bottomrule
		\end{tabular}
		\label{tab:hyperparameters}
	\end{table}
	
	\subsection{Models for Comparison}
	To comprehensively evaluate the performance of LCF-Net, we compare the LCA-Net with the following models.\\
	\textbf{LSTM}, \textbf{LCA-LSTM} We implement a baseline model for TSC based on the bi-directional LSTM (BiLSTM). Moreover, our implementation of LCA-LSTM is based on the LSTM, namely LCA-LSTM, in order to prove the effectiveness of the LCA framework.\\
	\textbf{IAN} \cite{ma2017interactive} employs interactive attentions to learn the semantic relatedness of context and targets. \\
	\textbf{RAM} \cite{chen2017recurrent} deploys multiple attentions to learn sentiment features combined with recurrent neural network and weighted-memory mechanism.\\
	\textbf{BiLSTM-ATT-G} \cite{liu2017attention} learns the features of left and right context using two attention-based LSTMs, and adjusts the contribution of left and right context features for polarity prediction according to gates.\\ 
	\textbf{MGAN} \cite{fan2018multi} adopts the fine-grained attention and coarse-grained attention mechanisms to learn the features of the context and the targets, and integrates a target-alignment loss to predict the polarity.\\
	\textbf{BERT-PT} \cite{xu2019bert} adapts the pretrained BERT to improve the performance of TSC based on post-training and fine-tuning.\\
	\textbf{AEN-BERT} \cite{song2019targeted} proposes an attentional encoder network based on the pretrained BERT to model for the context and the target. \\
	\textbf{BERT-SPC} \cite{devlin2019bert} is the pretrained BERT for sentence-pair classification which regards the context and the target as sentence-pair.\\
	\textbf{BERT-ADA} \cite{rietzler2019adapt} proposes the domain-adapted BERT for the Laptop and Restaurant datasets and obtains promising performance.\\
	\textbf{LCF-BERT} \cite{zeng2019lcf} employs the local context focus mechanism to extract local context features, which models for global context and local context using dual BERTs. We merge the dual BERTs in LCF-BERT to compare with LCA-BERT and analyze the effectiveness of LCE and LCP.\\
	\subsection{Overall Performance Analysis}
	
	\begin{table*}[t]
		\small
		\centering
		\caption{The experimental results (\%) of the LCA-Net. The results of comparative models are retrieved from the previous papers, and ``-'' denotes the not reported result. ``$\dagger$'' means the results are obtained by our implementations. ``$\ddagger$'' indicates the model that adopts domain-adapted BERT to improve performance.}
		\setlength{\tabcolsep}{2mm}{
			\begin{tabular}{cccccccc}
				\cmidrule{1-8}  & \multirow{2}[4]{*}{\textbf{Model}} & \multicolumn{2}{c}{\textbf{Laptop}} & \multicolumn{2}{c}{\textbf{Restaurant}} & \multicolumn{2}{c}{\textbf{Twitter}} \\
				\cmidrule(lr){3-4}\cmidrule(lr){5-6}\cmidrule(lr){7-8}  &       & Accuracy & macro F1 & Accuracy & macro F1 & Accuracy & macro F1 \\
				\midrule
				\multirow{2}[2]{*}{\textbf{LSTM models}} & LSTM$^\dagger$ & 70.22 & 64.36 & 77.50  & 67.17 & 69.49 & 67.64 \\
				& \textbf{LCA-LSTM} & 73.04 & 67.79 & 80.89 & 71.81 & 72.25 &70.05 \\
				\midrule
				\multirow{5}[2]{*}{\textbf{Baselines}} & TD-LSTM & 71.83 & 68.43 & 78.00    & 66.73 & 66.62 & 64.01 \\
				& BiLSTM-ATT-G& 73.12 & 69.80 & 79.73 & 69.25 & 70.38 & 68.37\\
				& RAM   & 74.49 & 71.35 & 80.23 & 70.80  & 69.36 & 67.3 \\
				& MGAN  & 75.39 & 72.47 & 81.25 & 71.94 & 72.54 & 70.81 \\
				& T-Net-LF & 76.01 & 71.47 & 80.79 & 70.84 & 74.68 & 73.36 \\
				& \textbf{LCA-MHSA} & 75.39 & 70.30  & 82.05 & 73.97 & 72.83 & 71.09 \\
				\midrule
				\multirow{7}[2]{*}{\textbf{BERT models}} & BERT-BASE$^\dagger$ & 79.00	& 75.59 & 82.59 & 75.36 & 74.13 & 72.19 \\
				& BERT-PT & 78.07 & 75.08 & 84.95 & 76.96 & -  & - \\
				& AEN-BERT & 79.93 & 76.31 & 83.12 & 73.76 & 74.71 & 73.13 \\
				& BERT-SPC$^\dagger$ & 80.09 & 76.39 & 85.62 & 78.94 & 75.58 & 74.35 \\
				& LCF-BERT$^\dagger$$^{\ddagger}$ & 80.72 & 78.05 & \textbf{89.11} & 83.86 & 75.72	& 74.34 \\
				& BERT-ADA$^\ddagger$ & 79.19 & 74.18 & 87.14 &80.05 & -  & -  \\
				&\textbf{LCA-BERT-E$^\ddagger$} &81.5 &77.36	&87.94	&82.07	&76.26	&75.64 \\
				&\textbf{LCA-BERT$^\ddagger$} &\textbf{82.45}&\textbf{79.22}	&88.93	&\textbf{83.96}	&\textbf{77.46}	&\textbf{76.17} \\
				
				\bottomrule
			\end{tabular}
		}
		\label{tab:mainresult}
	\end{table*}
	
	Table \ref{tab:mainresult} shows the main experimental results of the LCA-Net. Although LSTM is the basic neural network, the LSTM equipped with LCA techniques is competitive, which indicates that our framework is network-independent and easy to be integrated with other approaches. Compared with the Laptop and Restaurant datasets, all methods perform worse in the Twitter dataset. This is because there are lots of grammatical and spelling errors in the Twitter dataset. LCA-MHSA achieves the superior performance on the Restaurant dataset but underperforms T-Net in the Twitter dataset. The reason is that the convolutional neural network (CNN) is more competent to accurately extract features from ungrammatical sentences \cite{li2018transformation}. Compared to BERT-SPC and AEN-BERT, the experimental results indicate that the LCA-BERT obtains considerable performance on three datasets, especially the Laptop and Restaurant datasets (almost up to 3-4\%). Compared with BERT-PT and BERT-BASE, the performance of LCA-BERT on the Laptop dataset improved nearly 2-3\%. The LCF-BERT performs inferior to the LCA-Net on the Laptop and Twitter datasets. Experimental results show that LCA-BERT almost achieves state-of-the-art performance on the three datasets. With the same resource occupation and training time, the LCA-Net can achieve better results compared with other BERT-based models. The LCA-BERT-E achieves the better performance than existing methods, while the benchmark is behind the LCA-Net.

	\subsection{Ablation Experiments Analysis}
	
	We have listed the performance of LCA-LSTM, LCA-MHSA, and LCA-BERT in Table 3, proving that the LCA-Net improves the neural network-based methods. Next, we discuss the contributions of LCE, LCP, and CDM to the performance improvement through ablation analysis.
	Table \ref{tab:ablations} are experimental results of the ablated LCA-Net. It can be seen that the LCA-Net based on MHSA almost outperforms all the ablated models, which illustrates that the LCE and LCP proposed for LCA-BERT improves the model's capability of extracting local context features. According to our analysis, CDM contributes the most in the LCA-Net, followed by LCE and LCP. The performance of LCA-BERT on Laptop and Restaurant data machines is not as good as that of some ablated models (such as LCA-BERT w/o LCP and LCA-BERT w/o SPC), but it is worth noting that the training time of the LCA-Net is shorter and the convergence is faster compared to ablated the LCA-Net. LCA-BERT w/o ADA abandons the domain-adapted BERT to explore the performance of the LCA-Net which is based on BERT-BASE, and LCA-BERT w/o SPC removes the BERT-SPC trick. For the LCA-BERT, the LCA-BERT w/o SPC performs better than the baseline of LCA-BERT in the Laptop and Restaurant datasets (82.60\% and 89.38\% of accuracy, respectively), unexpectedly. Besides, the LCA-BERT outperforms all the ablated models in the Twitter dataset with a considerable accuracy of 77.46\%. In the absence of domain-adapted BERT, LCA-BERT is approximately 1-2\% ahead of other BERT-based models on the Laptop dataset.
	
	\begin{table*}[t]
		\small
		\centering
		\caption{The experimental results (\%) of ablated LCA-Nets. ``LCE'', ``LCP'' and ``CDM'' denote the local context embedding, local context tags prediction and local context focus, respectively. ``SPC'' and ``ADA'' refer to the BERT-SPC and domain-adapted BERT tricks, respectively.}
		\begin{tabular}{ccccccc}
			\toprule
			\multirow{2}[4]{*}{\textbf{Model}} & \multicolumn{2}{c}{\textbf{Laptop}} & \multicolumn{2}{c}{\textbf{Restaurant}} & \multicolumn{2}{c}{\textbf{Twitter}} \\
			\cmidrule{2-7}& Accuracy & macro F1 & Accuracy & macro F1 & Accuracy & macro F1 \\
			\midrule
			\textbf{LCA-MHSA} & \textbf{75.39} & \textbf{70.30}  & \textbf{82.05} & \textbf{73.97} & \textbf{72.83} & \textbf{71.09} \\
			w/o LCE & 72.10  & 67.38 & 81.16 & 71.91 & 71.53 & 70.49 \\
			w/o LCP & 74.29 & 70.06 & 81.34 & 73.19 & 72.25 & 70.53 \\
			w/o CDM & 73.51 & 68.49 & 79.73 & 69.57 & 70.81 & 68.79 \\
			\midrule
			\textbf{LCA-BERT} &82.45&79.20	&88.93	&83.96	&\textbf{77.46} &\textbf{76.17} \\
			w/o LCE &81.03 &77.88 &89.11 &84.52 &75.87 &73.90 \\
			w/o LCP &\textbf{82.60} &\textbf{79.42} &88.39 &83.07 &76.73 &74.10 \\
			w/o CDM &80.72 &77.45 &85.62 &77.81 &75.29 &73.84 \\
			w/o SPC &82.45 &79.12 &\textbf{89.38} &\textbf{84.66} &75.43 &73.91 \\
			w/o ADA &81.66 &78.63 &86.07 &79.12 &76.59 &75.48 \\
			\bottomrule
		\end{tabular}%
		\label{tab:ablations}%
	\end{table*}%

	\subsection{Discussion on Sigma ($\sigma$)}
\begin{figure}[h]
		\centering
		\subfigure[Accuracy]{
			\includegraphics[width=0.48\columnwidth]{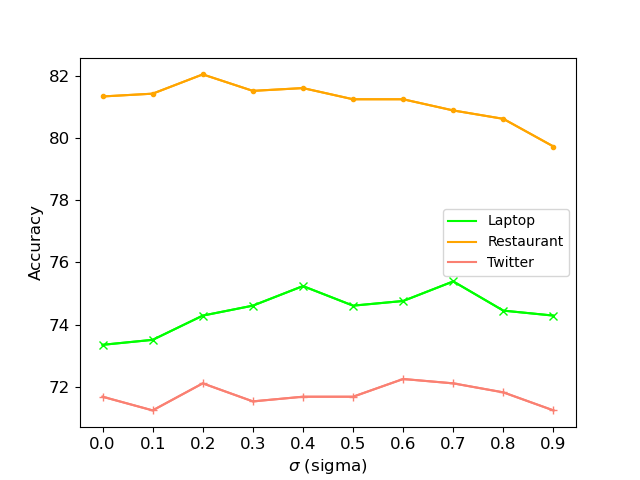}
		}
		\subfigure[macro F1]{
			\includegraphics[width=0.48\columnwidth]{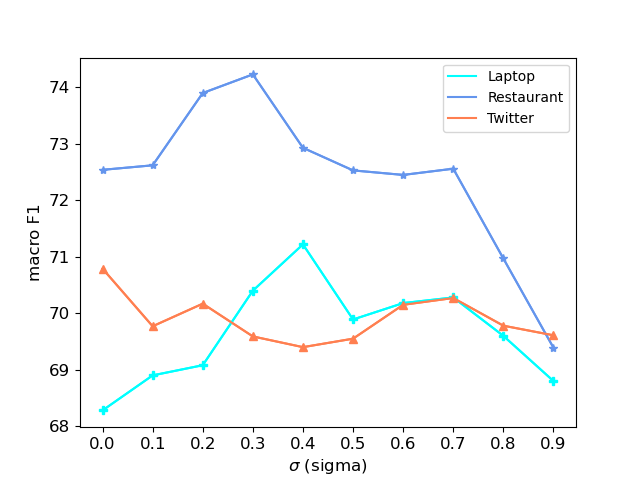}
		}
		\caption{The performance of LCA-MHSA on under different sigma($\sigma$).}
		\label{fig:sigma}
	\end{figure}

	The LCA-Net introduces an extra hyperparameter $\sigma$. To explore the influence of $\sigma$ on the performance, we set different values of $\sigma$ in the LCA-MHSA model to analyze $\sigma$'s impact. Fig. \ref{fig:sigma} are the experimental results on three datasets, which indicates the optimal $\sigma$ in the three datasets are different. The LCA-MHSA achieves better performance on the Laptop dataset when $\sigma\approx0.4$. The optimal effect on the Restaurant dataset was obtained when $\sigma\approx0.2$. For the Twitter dataset, the preferable performance was achieved when $\sigma = 0$ or $\sigma \approx 0.6$. Due to the limited computation resource, we do not conduct this experiment in LCA-BERT, and the $\sigma$ used by LCA-BERT in three datasets is 0.5.

	\subsection{Case Study}

	\begin{table*}[t]
		
		\centering
		\caption{Several cases of the TSC task. The words in \textbf{bold} and \textit{italic} are the targets and their local contexts ($\alpha=3$), respectively. The ``\checkmark'' and ``$\times$'' indicate the correct and error prediction of the LC-tags.}
		\setlength{\tabcolsep}{1.1mm}{
			\begin{tabular}{|c|r|c|}
				\hline
				No.& \multicolumn{1}{c|}{Sentences} & Polarity \\
				
				\hline
				\multirow{2}[4]{*}{1} & \multicolumn{1}{l|}{\textit{The \textbf{food} was extremely tasty}~~,~~creatively presented and the wine excellent .} & \multirow{2}[4]{*}{\textit{\textit{P}} \checkmark} \\
				\cline{2-2}          &\multicolumn{1}{l|}{~~\checkmark~~~~$\times$~~~~~\checkmark~~~~~~~~\checkmark~~~~~~~~\checkmark}  & \\
				
				\hline
				\multirow{2}[4]{*}{2} & \multicolumn{1}{l|}{It feels \textit{cheap~~,~~the \textbf{keyboard} is not very} sensitive .} & \multirow{2}[4]{*}{\textit{N} \checkmark} \\
				\cline{2-2}  & \multicolumn{1}{l|}{~~~~~~~~~~~~~~~~\checkmark~~~$\times$~~\checkmark~~~~~~~~\checkmark~~~~~~$\times$~~\checkmark~~~\checkmark}  & \\
				
				\hline
				\multirow{2}[4]{*}{3} & \multicolumn{1}{l|}{\textit{The \textbf{food} is surprisingly good~~,~~and the \textbf{decor} is nice .}} & \multirow{2}[4]{*}{\textit{P}\checkmark, \textit{P}\checkmark} \\
				\cline{2-2} & \multicolumn{1}{l|}{~~\checkmark~~~~\checkmark~~~\checkmark~~~~~~~~$\times$~~~~~~~~~~~~\checkmark~~~\checkmark~~$\times$~~~\checkmark~~~~~\checkmark~~~\checkmark~~\checkmark~~\checkmark}  & \\
				
				\hline
				\multirow{2}[4]{*}{4} & \multicolumn{1}{l|}{\textit{\textbf{Windows 7} can get 8} out of 10 viruses . Merry Christmas .} & \multirow{2}[4]{*}{\textit{P} $\times$ (\textit{O})} \\
				\cline{2-2} & \multicolumn{1}{l|}{~~~~~~~\checkmark~~~~~$\times$~~\checkmark~~~\checkmark~$\times$ ~~}  & \\
				
				\hline
				\multirow{2}[4]{*}{5} & \multicolumn{1}{l|}{I hope \textit{the songs of \textbf{britney spears} will be good} just like right now the next 20 years .} & \multirow{2}[4]{*}{\textit{P} \checkmark} \\
				\cline{2-2}          & \multicolumn{1}{l|}{~~~~~~~~~~~~$\times$~~~~~\checkmark~~~~$\times$~~~~~\checkmark~~~~~~~~\checkmark~~~~~~\checkmark~~~$\times$~~~\checkmark}  & \\
				\hline
			\end{tabular}%
		}
		\label{tab:casestudy}%
	\end{table*}%

	Table \ref{tab:casestudy} shows the predictions of some cases in the LCA-BERT. ``\textit{P}'', ``\textit{N}'', ``\textit{O}'' means positive, negative, and neutral, respectively. It can be seen that the local context-aware framework has a strong capability of target sentiment feature extraction. The LCA-Net achieves high accuracy in predicting local context tags. Consistent with other approaches, the LCA-Net performs slightly poor in predicting neutral sentiment compared to positive and neutral sentiment. More than half of the targets in the Twitter dataset contain neutral sentiment, which potentially results in low performance.
	According to the experimental results, the more LC-tags are correctly identified, the more likely the polarity of the aspect is to be accurately inferred, generally.
	
	\begin{figure*}[h]
		\centering
		\includegraphics[width=0.7\columnwidth]{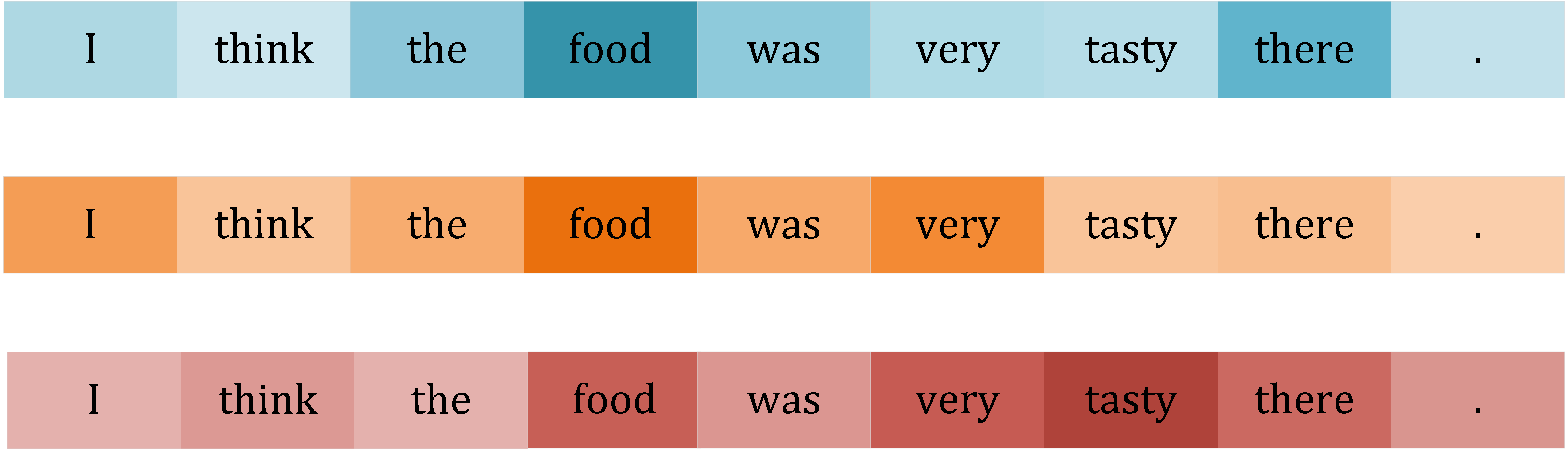}
		\caption{The visualization of output attention scores of several models. From top to bottom are the attention visualization of the LCF-BERT, the LCFS-BERT and the LCA-Net, respectively.}
		\label{fig:attention}
	\end{figure*}
	
	To evaluate the significance of LCE and LCP module, and to measure their impact on attention computation, we output and visualize the attention score of the models based on LCF and BERT, as shown in the Fig. \ref{fig:attention}. And the attention scores of all LCF-based models are sensitive to the target and its local context as expected.
	
	\section{Conclusion and Future Works}
	To exploit the significance of potential target-related information of local context, the local context embedding and local context prediction are proposed in this paper. Combining with the local context focus mechanism, we propose a novel framework called the LCA-Net which obtains state-of-the-art performance on three datasets. Besides, We conducted extensive ablation experiments to demonstrate the importance of LCE and LCP. To validate the transferability of the LCA-Net framework, we implement the LCA-Net based on LSTM, MHSA, and BERT, respectively, which indicates that it is a network-independent framework and can be easily adapted to other approaches. In the future, we will study the promotion of local context-aware techniques on other target-level NLP tasks, such as word sense disambiguation and part-of-speech tagging.
	
	\section*{Acknowledgments and Funding}
	Thanks to the anonymous reviewers and the scholars who helped us. 
	This research is supported by the Innovation Project of Graduate School of South China Normal University and funded by National Natural Science Foundation of China, Multi-modal Brain-Computer Interface and Its Application in Patients with Consciousness Disorder, Project approval number: 61876067.

	\bibliographystyle{coling}
	\bibliography{coling2020}

\end{document}